\documentclass[preprint,12pt]{elsarticle}

\usepackage{colortbl}
\usepackage{booktabs}
\usepackage{array}
\usepackage{colortbl}
\usepackage{multirow}
\usepackage{arydshln}
\usepackage{caption}
\usepackage{pifont}
\usepackage{graphicx}
\usepackage{subfig}
\usepackage{algorithm}
\usepackage{algorithmic}
\usepackage{listings}
\usepackage{multirow}
\usepackage[table]{xcolor}
\usepackage{arydshln}
\usepackage{tabularx}
\usepackage{pifont}
\usepackage{amsmath}
\usepackage{amssymb}
\usepackage{hyperref}

\hypersetup{colorlinks = true, linkcolor = blue, anchorcolor = red, citecolor = blue, filecolor = red, urlcolor = red,
            pdfauthor=author}
\newcommand{\ie}{\textit{i}.\textit{e}., }

\makeatletter
\def\ps@pprintTitle{%
 \let\@oddhead\@empty
 \let\@evenhead\@empty
 \let\@oddfoot\@empty
 \let\@evenfoot\@empty
}
\makeatother

\begin{document}

\begin{frontmatter}

\title{IAM: Enhancing RGB-D Instance Segmentation \\ with New Benchmarks}

\author[1]{Aecheon~Jung\fnref{equal}}
\ead{kasurashan@skku.edu}

\author[1]{Soyun~Choi\fnref{equal}}
\ead{sychoi9719@skku.edu}

\author[2]{Junhong~Min}
\ead{junhong.min.85@gmail.com}

\author[1]{Sungeun Hong\corref{cor1}}
\cortext[cor1]{Corresponding author}
\ead{csehong@skku.edu}

\fntext[equal]{These authors contributed equally to this work.}

\affiliation[1]{
            addressline={Department of Immersive Media Engineering}, organization={Sungkyunkwan University}}
\affiliation[2]{            addressline={Global Technology Research}, organization={Samsung Electronics}}
\begin{abstract}

{Image segmentation is a vital task for providing human assistance and enhancing autonomy in our daily lives. 
In particular, RGB-D segmentation—leveraging both visual and depth cues—has attracted increasing attention as it promises richer scene understanding than RGB-only methods. However, most existing efforts have primarily focused on semantic segmentation and thus leave a critical gap. There is a relative scarcity of instance-level RGB-D segmentation datasets, which restricts current methods to broad category distinctions rather than fully capturing the fine-grained details required for recognizing individual objects.
To bridge this gap, we introduce three RGB-D instance segmentation benchmarks, distinguished at the instance level.
These datasets are versatile, supporting a wide range of applications from indoor navigation to robotic manipulation.
In addition, we present an extensive evaluation of various baseline models on these benchmarks. This comprehensive analysis identifies both their strengths and shortcomings, guiding future work toward more robust, generalizable solutions. 
Finally, we propose a simple yet effective method for RGB-D data integration. Extensive evaluations affirm the effectiveness of our approach, offering a robust framework for advancing toward more nuanced scene understanding.}

\end{abstract}


\end{frontmatter}

\section{Introduction}

The human visual system reconstructs 3D scenes from 2D images using depth cues such as binocular disparity and texture gradients~\cite{tsutsui2005neural}. Inspired by this, incorporating depth information has significantly advanced segmentation research, addressing the limitations of traditional RGB-only methods~\cite{qi20173d,zhang2022cmx,chen2020bi}.
One of the key advantages of depth information lies in its ability to enhance object boundaries using geometric cues. Traditional RGB segmentation models often rely on post-processing~\cite{tang2021look} or explicit boundary learning~\cite{kim2021devil}. However, they frequently struggle under low illumination or when objects share similar textures with the background. These challenges particularly impact tasks such as object detection and image segmentation, which require precise boundary delineation. Depth information mitigates these issues by enriching boundary details and capturing geometric features that RGB data may miss~\cite{wang2018depth}.

Despite its advantages, existing RGB-D research has primarily focused on semantic segmentation~\cite{wang2018depth,cao2021shapeconv,zhang2022cmx}, which classifies pixels into predefined categories but does not differentiate between individual object instances. To overcome this limitation, RGB-D instance segmentation is necessary. It enables the recognition and separation of individual objects, even when they belong to the same category. For example, in autonomous driving, accurately detecting and distinguishing vehicles, pedestrians, and obstacles is critical for safe navigation~\cite{zhang2016instance,de2017semantic_autonomous}. Similarly, in robotics, instance segmentation allows robots to differentiate and manipulate objects, facilitating smoother human-robot and robot-robot interactions~\cite{back2022unseen,ito2020point}.

Advancing RGB-D segmentation to instance segmentation represents a natural and necessary progression. However, one major challenge is the lack of benchmark datasets specifically designed for RGB-D instance segmentation. Table~\ref{table:1} highlights this issue. While comprehensive datasets exist for RGB and RGB-D semantic segmentation, including those for indoor environments, large-scale RGB-D datasets for instance segmentation remain limited and often small in size. Furthermore, many existing datasets rely heavily on RGB data or synthetic images, which reduces their applicability to real-world scenarios. This lack of dedicated datasets has slowed progress in RGB-D instance segmentation research. This is particularly significant given its potential for practical applications, such as indoor navigation and robotic tasks like box depalletizing, where precise object recognition and instance-level understanding are critical.

In this study, we address these limitations by introducing three RGB-D instance segmentation benchmarks specifically designed to capture the complexity and diversity of real-world indoor environments. To construct these benchmarks, we carefully refine and extend two widely recognized RGB-D datasets—NYUDv2~\cite{silberman2012indoor} and SUN-RGBD~\cite{song2015sun}—by re-annotating and reorganizing them for instance segmentation tasks. Additionally, we present a newly developed RGB-D Box dataset, designed to support applications requiring high-precision object recognition, such as human-robot interactions and box depalletizing. Each dataset includes thorough documentation and statistical analysis, ensuring it serves as a reliable resource for advancing RGB-D instance segmentation research.

To address the challenges of multimodal data fusion, we propose the Intra-modal Attention Mix (IAM) module, a novel and flexible approach for RGB-D instance segmentation. Unlike existing methods that focus heavily on inter-modal fusion, IAM enhances intra-modal feature relationships while efficiently integrating complementary information across RGB and depth modalities. By leveraging a mixup-inspired strategy~\cite{zhang2018mixup}, IAM balances modality-specific feature refinement and cross-modal connections. This mitigates common issues such as modality imbalance and over-reliance on single data types. The IAM module generates enhanced attention maps within each modality while minimizing redundant feature interactions, preserving critical spatial and geometric information necessary for accurate segmentation. This design boosts segmentation accuracy and reduces computational overhead compared to conventional fusion strategies.

\begin{figure}[!t]
\centering
\subfloat[\footnotesize Input\label{fig:Figure1_1}]{\includegraphics[width=.2127\columnwidth]{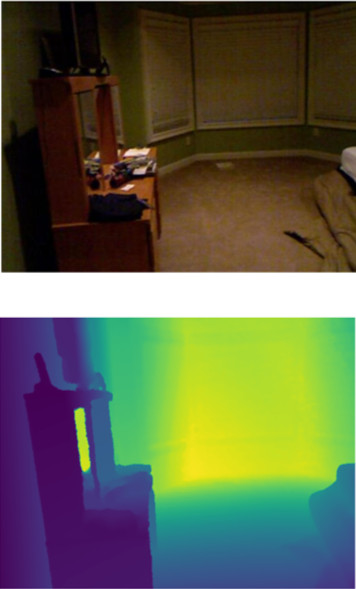}}
\hspace{2mm}
\subfloat[\footnotesize RGB\label{fig:Figure1_2}]{\includegraphics[width=.2147\columnwidth]{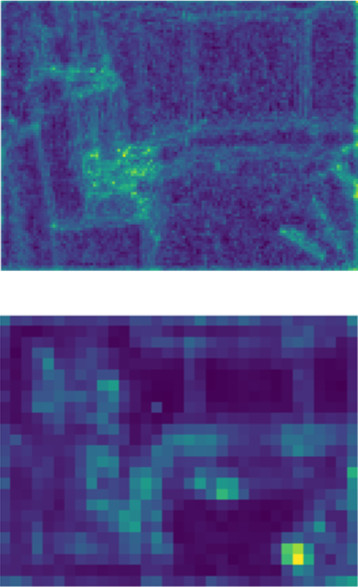}}
\hspace{2mm}
\subfloat[\footnotesize Depth \label{fig:Figure1_3}]{\includegraphics[width=.2146\columnwidth]{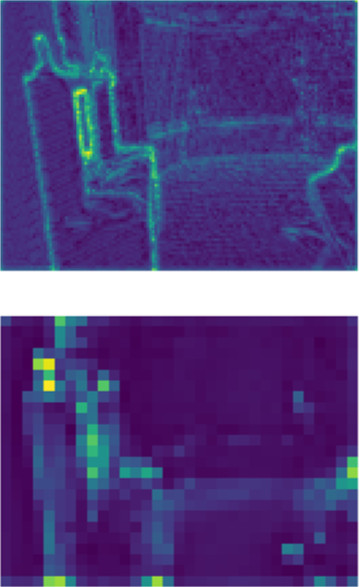}}
\hspace{2mm}
\subfloat[\footnotesize RGB-D\label{fig:Figure1_4}]{\includegraphics[width=.2178\columnwidth]{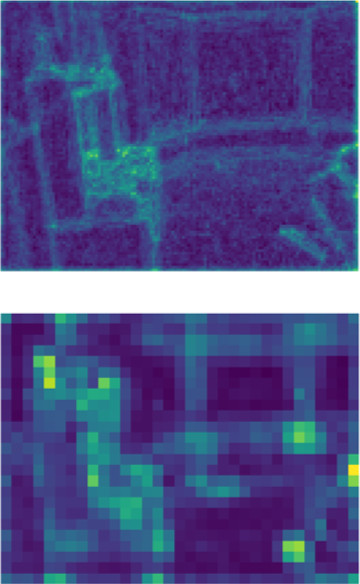}}
\vspace{-2mm}
\caption{Feature visualizations from the input image and depth map (a), as well as RGB and RGB-D models. (b) RGB features, (c) depth features, and (d) RGB-D features are shown. The first row presents fine-grained features, while the second row highlights progressively coarser features.}
\label{fig:Figure1}
\end{figure}

In our preliminary study, we extended a conventional RGB instance segmentation model into a two-branch RGB-D framework to validate the importance of depth information. This framework processes RGB and depth data through separate backbones and integrates their features using a fusion module placed between backbone layers. As shown in Figure~\ref{fig:Figure1_2}, features from the RGB backbone can be noisy, particularly under low-light conditions or when objects share similar textures. This makes it difficult to capture object boundaries. In contrast, Figure~\ref{fig:Figure1_3} demonstrates that depth features effectively delineate boundaries, even under challenging conditions.
Experimental results on our three benchmarks demonstrate that IAM consistently outperforms existing fusion methods, including early fusion, late fusion, and attention-based approaches~\cite{hu2019acnet,zhou2020rgb}. Notably, IAM achieves significant improvements in boundary delineation and instance separation, which are critical for accurate RGB-D instance segmentation. These results highlight IAM’s capability to deliver fine-grained object recognition efficiently, making it a robust and scalable solution for RGB-D instance segmentation tasks.

In summary, our contributions are as follows:
\begin{itemize} 
\item We construct two public RGB-D instance segmentation dataset by carefully refining and extending widely used RGB-D semantic datasets with instance-level annotations. Additionally, we introduce the RGB-D Box dataset, specifically designed for precise object recognition in tasks such as box depalletizing and human-robot interaction.

\item We propose the Intra-modal Attention Mix (IAM) module, a novel and lightweight module that improves RGB-D instance segmentation by enhancing intra-modal interactions and integrating inter-modal features efficiently. 

\item Extensive experiments validate IAM's effectiveness, showing consistent improvements over existing fusion methods. On NYUDv2-IS, our proposed method achieves a 2.8\% gain over intra-modal attention and 6.7\% over early fusion, with similar results on SUN-RGBD-IS and Box-IS. These findings establish IAM as a strong baseline in the emerging field of RGB-D instance segmentation, enabling future advancements.
\end{itemize}

\section{Related Work}

\subsection{RGB-D Semantic Segmentation}

RGB-D semantic segmentation integrates RGB and depth data to classify each pixel into semantic categories. This approach addresses limitations of traditional RGB-only methods, such as low-light conditions, texture ambiguities, and distinguishing objects with similar colors. By leveraging geometric cues from depth information, it enhances scene understanding and segmentation accuracy in complex environments.
Early methods, like RDFNet~\cite{park2017rdfnet}, utilized residual learning to fuse RGB and depth features effectively, employing skip connections to preserve fine details. Depth-aware CNN~\cite{wang2018depth} introduced depth-aware convolution operations that integrate geometric information for improved boundary localization. Later, ShapeConv~\cite{cao2021shapeconv} decomposed image patches into base and shape components for shape-aware feature processing. UCTNet~\cite{ying2022uctnet}, inspired by Vision Transformers~\cite{dosovitskiy2020image}, modeled long-range dependencies for efficient feature fusion. Recently, techniques like DPLNet~\cite{dong2023efficient} leveraged prompt-learning to reduce trainable parameters while maintaining performance.

Despite its progress, RGB-D semantic segmentation faces challenges such as handling noisy or incomplete depth data and high computational costs of multi-modal fusion. Moreover, the pixel-level nature of this task struggles with overlapping or occluded objects. To address these limitations, RGB-D instance segmentation has emerged, focusing on object-level segmentation and leveraging the strengths of RGB and depth data.

\subsection{Instance Segmentation}

Instance segmentation methods are broadly classified into two-stage, multi-stage, and single-stage approaches. Two-stage methods include top-down and bottom-up strategies. Top-down approaches~\cite{he2017mask,wang2019panet} detect instance bounding boxes first and segment contents within each box, while bottom-up approaches~\cite{liang2017proposal,huang2019mask} generate instance masks directly using clustering techniques. While effective, bottom-up methods often involve computationally expensive post-processing steps. Multi-stage methods~\cite{cai2019cascade,chen2019hybrid} refine segmentation results across successive stages, enhancing accuracy but increasing computational demand.
Single-stage methods~\cite{bolya2019yolact,tian2020conditional} integrate detection and segmentation into a unified network, optimizing both tasks for greater efficiency. Transformer-based models~\cite{carion2020end,dong2021solq}, inspired by Vision Transformers~\cite{DBLP:conf/iclr/DosovitskiyB0WZ21}, introduced bipartite matching for set prediction, eliminating the need for post-processing steps like Non-Maximum Suppression (NMS).

RGB-D instance segmentation remains underexplored due to the scarcity of high-quality datasets. While extensive datasets exist for RGB-based tasks, RGB-D datasets often rely on synthetic data~\cite{hou2021pri3d,wang2023amnet}. To address this gap, we developed NYUDv2-IS and SUN-RGBD-IS by refining existing semantic segmentation datasets for instance-level tasks. Additionally, the Box-IS dataset was created to support human-robot interaction tasks, such as box depalletizing. These benchmarks aim to advance research in precise and reliable RGB-D object segmentation.

\subsection{RGB-D Fusion Methods}

RGB-D fusion methods have been extensively explored in semantic segmentation tasks. Early approaches, such as FuseNet\cite{hazirbas2016fusenet} and RedNet\cite{jiang2018rednet}, performed simple element-wise addition to combine RGB and depth features. While computationally efficient, these methods fail to capture the complementary nature of RGB and depth information effectively.

Recent works have proposed more advanced multi-modal fusion strategies to enhance feature integration. ACNet\cite{hu2019acnet} introduced an attention mechanism to selectively fuse RGB and depth features based on their information content, while NaNet\cite{zhang2021non} employed a non-local aggregation module for improved multi-modal interactions. Methods like MGCNet\cite{yang2022mgcnet} addressed modality differences using a difference-exploitation fusion module and gated decoders, and CMX\cite{zhang2022cmx} further explored spatial and channel-wise interactions to maximize the utility of complementary information. Similarly, PGDENet\cite{zhou2022pgdenet} introduced a progressive guided fusion strategy and depth enhancement techniques, while CAINet\cite{lv2024context} leveraged global contextual relationships for RGB-D tasks, designed for RGB-T semantic segmentation.

Despite their advancements, existing approaches primarily focus on inter-modal feature fusion while overlooking transformations within individual modalities. This can lead to suboptimal use of modality-specific information. To address this, we propose a novel Intra-modal Attention Mix (IAM) module that enhances feature relationships within each modality while efficiently integrating cross-modal information. By balancing intra- and inter-modal attention, our method reduces computational overhead and preserves critical spatial and geometric details, achieving superior performance compared to existing fusion methods.

\setlength{\tabcolsep}{0.2cm}
\begin{table*}[]
\centering

\footnotesize
\resizebox{\columnwidth}{!}{
\begin{tabular}{clccrrrrr}
\toprule
\multicolumn{1}{c}{Task Type} & \multicolumn{1}{c}{Dataset} & \multicolumn{1}{c}{RGB} & \multicolumn{1}{c}{Depth} & \multicolumn{1}{c}{Images} & \multicolumn{1}{c}{Resolution} & \multicolumn{1}{c}{Classes} & \multicolumn{1}{c}{Scene Type} & \multicolumn{1}{c}{Depth Sensor Type} \\
\midrule
\multirow{5}{*}{Semantic} & COCO-stuff~\cite{caesar2018coco} & \checkmark & - & 163,957 & Variable & 172 & Indoor, Outdoor & - \\
& ADE20K~\cite{zhou2017scene} & \checkmark & - & 25,210 & Variable & 150 & Indoor, Outdoor & - \\
& NYUDv2~\cite{silberman2012indoor} & \checkmark & \checkmark & 1,449 & $640 \times 480$ &  13/40 & Indoor & Kinect v1 \\
& SUN-RGBD~\cite{song2015sun} & \checkmark  & \checkmark & 10,335 & $730 \times 530$ & 37 & Indoor & Intel RealSense, Asus Xtion, Kinect v1/v2 \\
& Cityscapes~\cite{cordts2016cityscapes} & \checkmark & \checkmark & 25,000 & $2,048 \times 1,024$ &  30 & Outdoor & Stereo Camera \\
\midrule
\multirow{10}{*}{Instance} & COCO~\cite{lin2014microsoft} & \checkmark & - & 123,287 & Variable & 80 & Indoor, Outdoor & - \\
& Hypersim~\cite{roberts2021hypersim} & \checkmark & \checkmark & 77,400 & $1024 \times 768$ & 40 & Indoor & Synthetic \\
& SYNTHIA~\cite{ros2016synthia} & \checkmark & \checkmark & 200,000 & $960 \times 720$ & 13 & Outdoor & Synthetic \\
& Virtual Kitti\cite{gaidon2016virtual} & \checkmark & \checkmark & 21,260 &  $1242 \times 375$ & 14 & Outdoor & Synthetic \\
& Woodscape~\cite{yogamani2019woodscape}&\checkmark & \checkmark & 10,000 & $1280 \times 1024 $ & 40 & Outdoor & LiDAR \\
& OSD~\cite{richtsfeld2012segmentation} & \checkmark & \checkmark & 111 & $640 \times 480$ & 1 & Objects & Kinect-style \\
& YCB~\cite{calli2015ycb} & \checkmark & \checkmark & 600 & $1280 \times 1024$ & 77 & Objects & Kinect-style \\
\cmidrule{2-9}
\multicolumn{1}{l}{} & NYUDv2-IS & \checkmark & \checkmark & 1,433 & $640 \times 480$ &  9 & Indoor & Kinect v1 \\
& SUN-RGBD-IS & \checkmark  & \checkmark & 9,942 & $730 \times 530$ & 17 & Indoor & Intel RealSense, Asus Xtion, Kinect v1/v2 \\
& Box-IS & \checkmark &\checkmark & 543 & $1280 \times 720$ & 1 & Indoor & Intel RealSense\\
\bottomrule
\end{tabular}
}
\caption{Specifications of image segmentation datasets.}
\label{table:1}
\end{table*}
\setlength{\tabcolsep}{0.2cm}

\section{Datasets}

\begin{figure}[!t]
\centering

\subfloat[\footnotesize NYUDv2-IS\label{fig:Figure2_1}
]{\includegraphics[width=0.75\columnwidth]{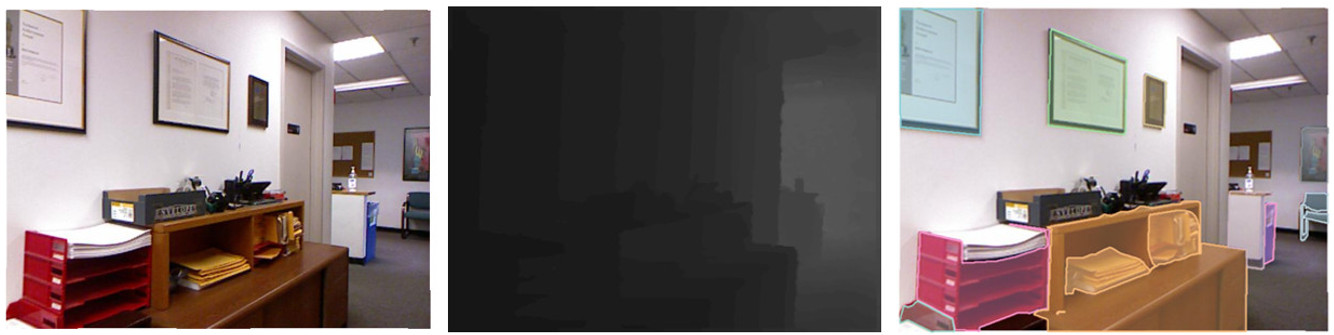}}
\vspace{-2.5mm}

\subfloat[\footnotesize SUN-RGBD-IS\label{fig:Figure2_2}
]{\includegraphics[width=0.75\columnwidth]{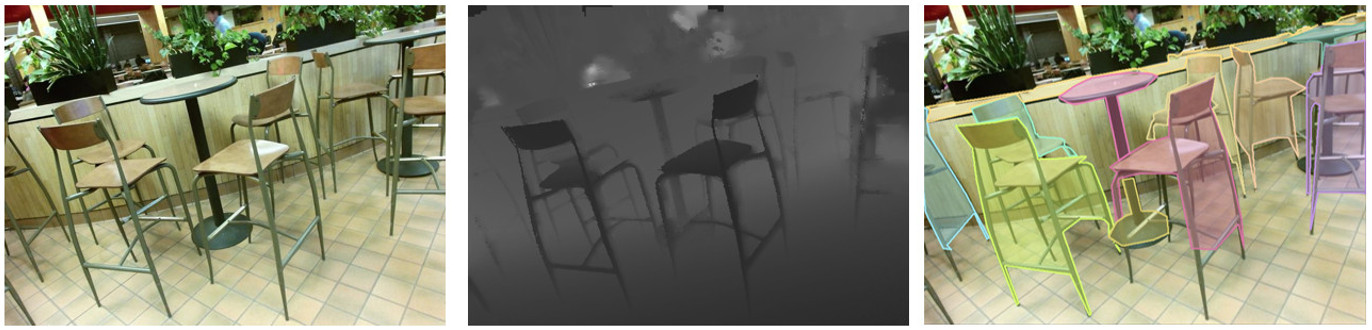}}
\vspace{-2.5mm}

\subfloat[\footnotesize Box-IS\label{fig:Figure2_3}
]{\includegraphics[width=0.75\columnwidth]{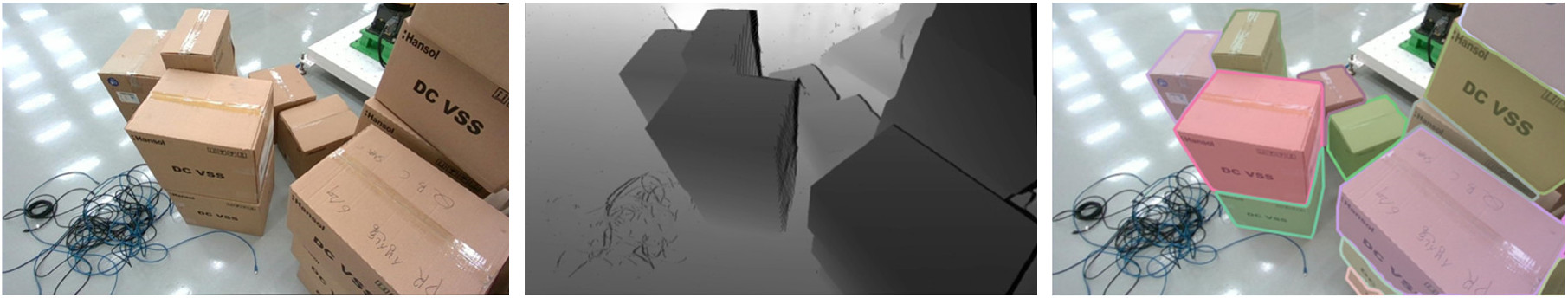}}
\vspace{-2.5mm}

\caption{Examples of datasets: RGB image, depth map, and ground truth labels.}
\label{fig:Figure2}
\end{figure}

{
RGB-D instance segmentation faces a significant challenge due to the lack of dedicated benchmark datasets, as shown in Table~\ref{table:1}.
In this section, we introduce the datasets utilized in our study, specifically NYUDv2-IS, SUN-RGBD-IS, and Box-IS, as depicted in Figure~\ref{fig:Figure2}. We describe their construction, unique characteristics, and the steps taken to adapt them for instance segmentation benchmarks.}

\subsection{Dataset Construction}
NYUDv2~\cite{silberman2012indoor} is a widely-used RGB-D benchmark dataset for semantic segmentation, focusing on real-world indoor scenes. Depth data was captured using the Kinect v1 sensor, which provides reliable depth information within indoor environments. To construct NYUDv2-IS, specifically tailored for instance segmentation, we generated instance masks that delineate individual objects in each image. These masks were labeled using the object class annotations provided in the original NYUDv2 dataset, which is distributed in MATLAB format.
The process involved several key steps: (1) extracting binary instance masks, (2) converting these masks into polygon representations, and (3) generating COCO-style annotations. Each annotation includes essential attributes such as category ID, segmentation masks, bounding boxes, object areas, and image metadata. During this conversion, we focused on 9 categories out of the original 13 classes, excluding non-instance categories such as walls and floors. To ensure dataset quality, images without any object annotations were systematically removed.
This benchmark dataset is publicly accessible at \href{https://github.com/AIM-SKKU/NYUDv2-IS}{https://github.com/AIM-SKKU/NYUDv2-IS}.

SUN-RGBD~\cite{song2015sun} is a large-scale RGB-D dataset extensively used for semantic segmentation tasks. It includes diverse indoor environments, such as homes, classrooms, offices, and retail spaces, with depth data collected using four sensor types: Intel Realsense, Asus Xtion, Kinect v1, and Kinect v2. This variety ensures robustness across sensors and settings.
To transform SUN-RGBD into an instance segmentation benchmark (\ie SUN-RGBD-IS), we employed a pipeline similar to that of NYUDv2-IS. We selected 17 categories from the original 37 classes, carefully omitting non-instance categories like ceilings and walls. Images lacking any identifiable object instances were filtered out to maintain dataset relevance for instance segmentation tasks.
We systematically convert segmentation annotations into COCO format, generating precise bounding boxes, instance masks, and object attributes. SUN-RGBD-IS dataset is available at \href{https://github.com/AIM-SKKU/SUN-RGBD-IS}{https://github.com/AIM-SKKU/SUN-RGBD-IS}.

The Box-IS dataset was created to support research on human-robot collaboration with a focus on robotic manipulation tasks. 
It was captured using the Intel\textsuperscript{\textregistered} RealSense\texttrademark{} Depth Camera D455, a high-performance sensor designed for depth imaging. 
To ensure precise depth measurements, we bypassed the default depth data processing of the sensor and performed accurate stereo matching directly from the captured left and right IR images. Employing the UniMatch technique~\cite{xu2023unifying}, we derived a high-quality depth map from these stereo IR images, which was then aligned with the corresponding RGB image for a comprehensive output.
The dataset was intentionally designed to encompass a broad range of scene complexities, from simple box arrangements to highly irregular configurations. This diversity ensures that it can effectively benchmark algorithms across varying levels of difficulty.
Our Box-IS benchmark dataset can be accessed at \href{https://github.com/AIM-SKKU/Box-IS}{https://github.com/AIM-SKKU/Box-IS}.

\newcolumntype{R}[1]{>{\raggedleft\arraybackslash}p{#1}} 
\setlength{\tabcolsep}{0.2cm}
\begin{table}[!t]
\begin{center}

\normalsize
\resizebox{0.8\columnwidth}{!}{
\begin{tabular}{lrrR{1.8cm}R{1.8cm}}
\toprule
\multicolumn{1}{c}{Dataset} & \multicolumn{1}{c}{Images} & \multicolumn{1}{c}{Classes} & \multicolumn{1}{c}{\# of objects / img}& \multicolumn{1}{c}{\# of categories / img}  \\
\midrule
NYUDv2-IS & 1,433 & 9 &  6.4 & 3.4 \\
SUN-RGBD-IS & 9,942 & 17 & 5.1 & 2.6 \\
Box-IS & 543 & 1 & 12.3 & 1 \\
\bottomrule
\end{tabular}
}
\end{center}
\vspace{-3.5mm}
\caption{Comparison of instance segmentation datasets.}
\label{table:2}
\end{table}
\setlength{\tabcolsep}{0.2cm}

\begin{figure*}[]
\vspace{3.5mm}
\centering
\hspace{2 mm}
\subfloat[\footnotesize Diversity of categories per image\label{fig:Figure3_1}]{\includegraphics[width=.3\columnwidth]{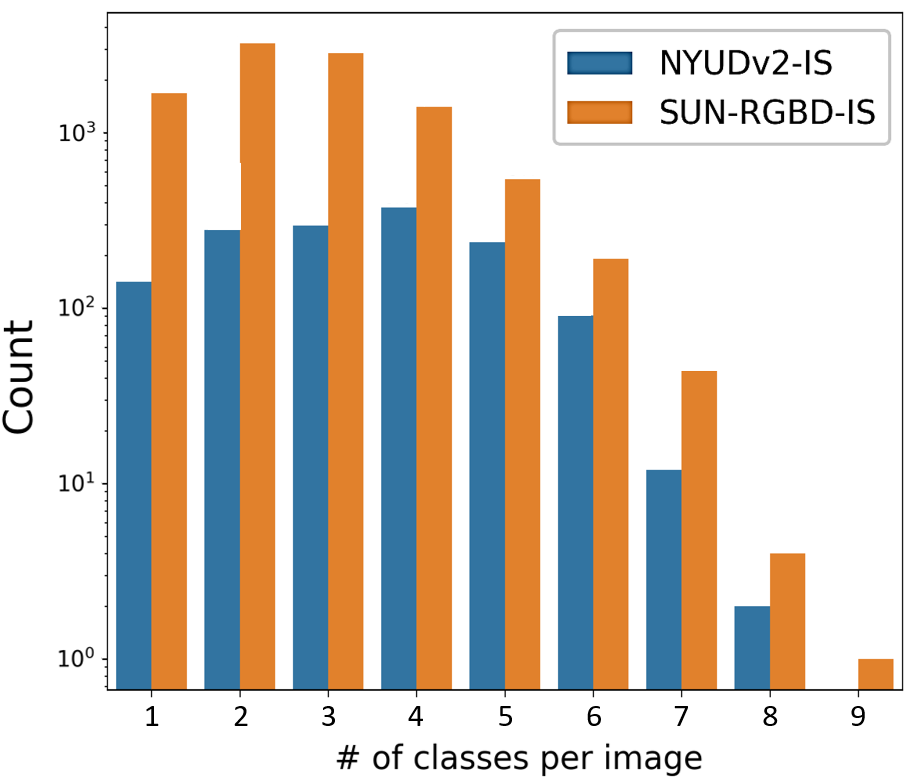}}
\hspace{9 mm}
\subfloat[\footnotesize The number of instances per category\label{fig:Figure3_2}]{\includegraphics[width=0.60\columnwidth]{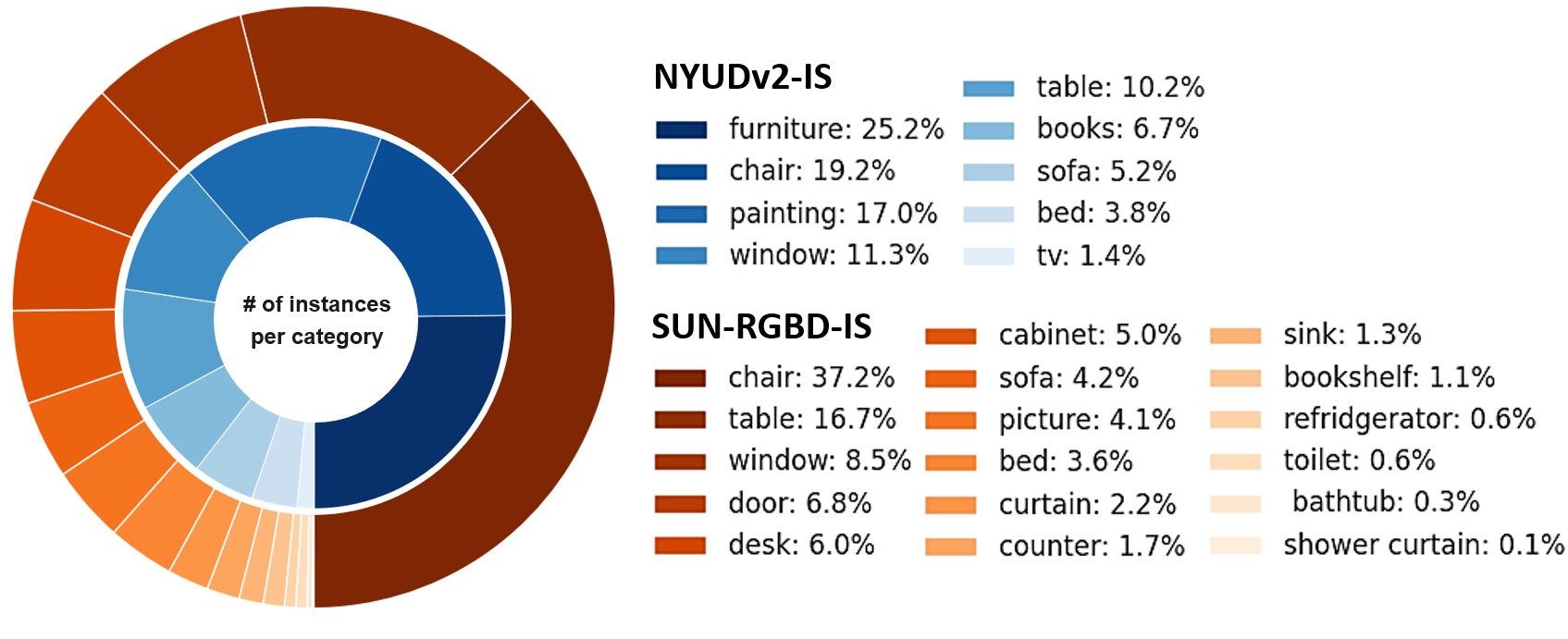}}
\vspace{-1.5mm}
\caption{Examples from the constructed datasets. (a) Diversity of categories per image, comparing the number of classes per image. (b) Distribution of instances per category, illustrating the proportions of object categories.}
\label{fig:Figure3}
\end{figure*}

\subsection{Dataset Statistics}

Table~\ref{table:2} summarizes the average number of objects and classes per image for each dataset. NYUDv2-IS and SUN-RGBD-IS contain relatively fewer objects and classes per image. This limitation arises from the nature of indoor environments, where object diversity is lower compared to outdoor scenes. On the other hand, the Box-IS dataset includes a much higher number of objects per image. However, since it focuses exclusively on box arrangements, it consists of only one class, reducing overall category diversity.

Figure~\ref{fig:Figure3_1} shows the distribution of categories per image. SUN-RGBD-IS generally contains fewer categories compared to NYUDv2-IS, reflecting a lower diversity of objects within individual scenes. Figure~\ref{fig:Figure3_2} highlights the class imbalance across all datasets. In both NYUDv2-IS and SUN-RGBD-IS, certain classes dominate while others have significantly fewer samples. This imbalance creates challenges for training instance segmentation models, as models tend to favor frequently occurring classes while underperforming on less-represented ones.

\begin{figure}[!t]
\centering
\includegraphics[width=.50\columnwidth]{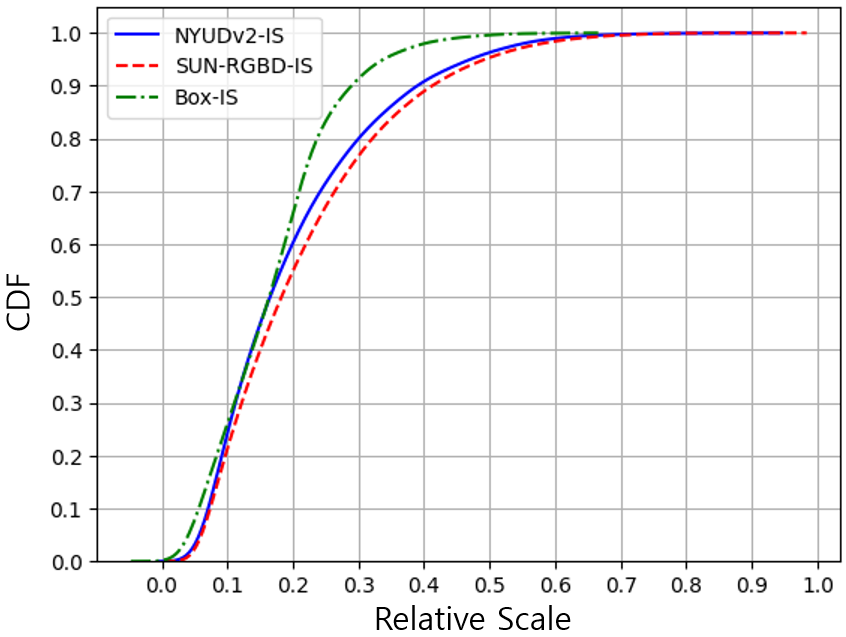}
\caption{Cumulative distribution function showing the relative scale of objects.}
\label{fig:Figure10}
\end{figure}

Figure~\ref{fig:Figure10} shows the cumulative distribution function (CDF) of the relative scale of objects, defined as the square root of the ratio between the object’s segmentation mask area and the total image area. Across all datasets, over 50\% of objects occupy less than 4\% of the image area, compared to the COCO dataset, where most objects take up less than 1\%~\cite{singh2018analysis}. This highlights a lower prevalence of small objects. NYUDv2-IS and SUN-RGBD-IS share similar distributions, with about 90\% of objects covering less than 16\% of the area. However, the Box-IS dataset shows a higher concentration of smaller objects, with 90\% occupying less than 9\%, reflecting its focus on tasks like box depalletizing.

\begin{figure}[!t]
\centering
\subfloat[\footnotesize NYUDv2-IS\label{fig:Figure11_1}]{\includegraphics[width=.31\columnwidth]{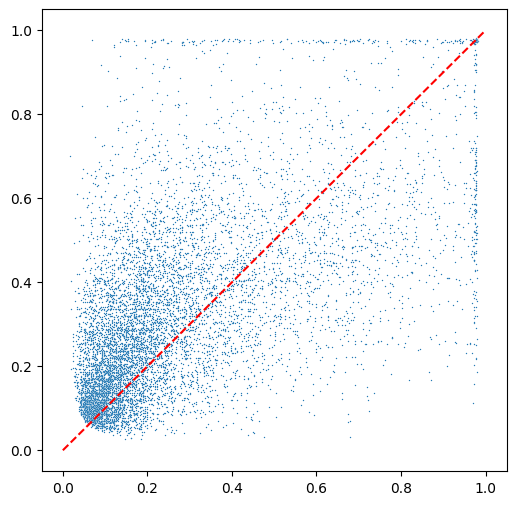}}
\hspace{3mm}
\subfloat[\footnotesize SUN-RGBD-IS\label{fig:Figure11_2}]{\includegraphics[width=.31\columnwidth]{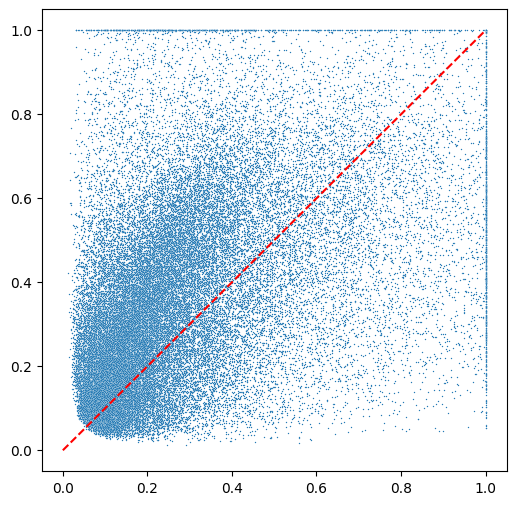}}
\hspace{3mm}
\subfloat[\footnotesize Box-IS\label{fig:Figure11_3}]{\includegraphics[width=.31\columnwidth]{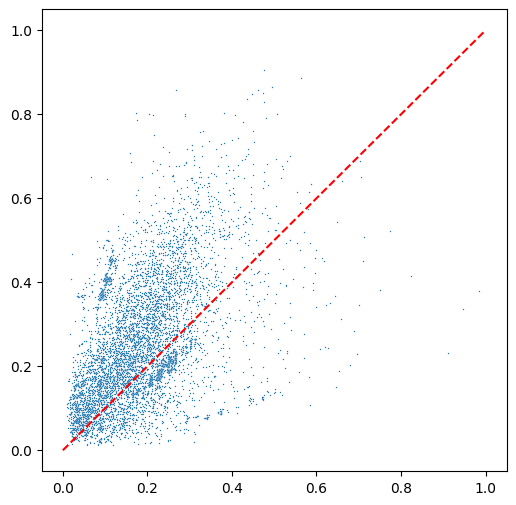}}
\caption{Scatter plots illustrating the relative proportions of bounding boxes for each instance within an image. The x-axis represents the relative width of the bounding boxes, while the y-axis represents the relative height. Each point on the plot corresponds to an instance, showcasing its unique bounding box proportions.}
\label{fig:Figure11}
\end{figure}

Figure~\ref{fig:Figure11} presents a scatter plot of bounding box aspect ratios and relative sizes. Points near the red dashed line indicate square shapes, while points in the upper-right represent larger bounding boxes. Box-IS primarily clusters near the red line, showing smaller bounding boxes and fewer extreme aspect ratios. NYUDv2-IS and SUN-RGBD-IS display greater variation in aspect ratios and object sizes, with SUN-RGBD-IS showing particularly diverse patterns. Large objects like beds and sofas in indoor scenes cluster near $x=1.0$ and $y=1.0$. Tilted objects result in oversized bounding boxes, highlighting challenges with object orientation and dataset characteristics that can affect real-world model performance.

\section{Method}

\subsection{Baseline Models}

\begin{figure}[!t]
\centering
\captionsetup{labelfont=tiny, textfont=tiny}
    \subfloat[Early fusion]
    {\includegraphics[width=0.18\linewidth]{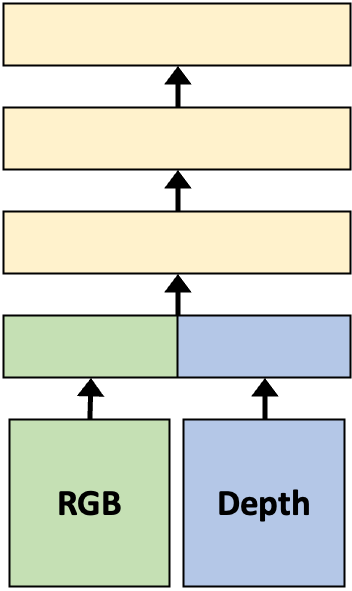}}\hfill \hfill \hfill
    \subfloat[Late fusion]
    {\includegraphics[width=0.1826\linewidth]{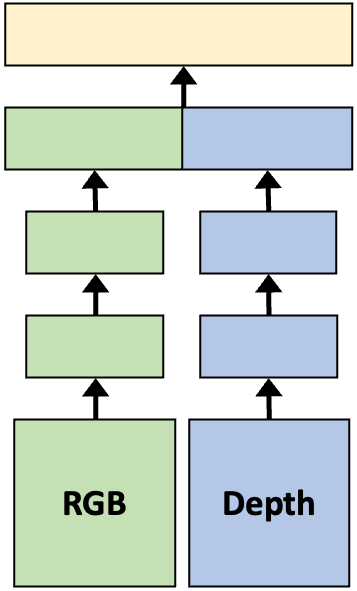}}\hfill
    \subfloat[Intra-modal]
    {\includegraphics[width=0.199\linewidth]{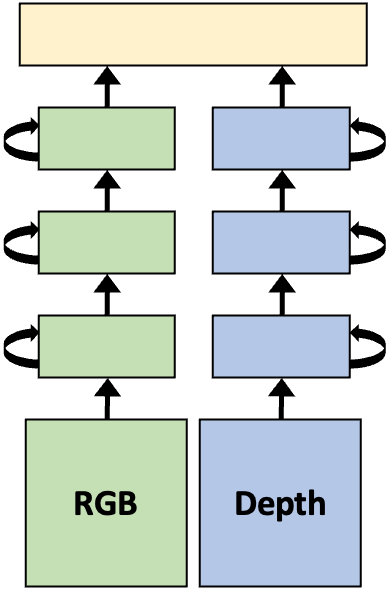}}\hfill
    \subfloat[Inter-modal]
    {\includegraphics[width=0.192\linewidth]{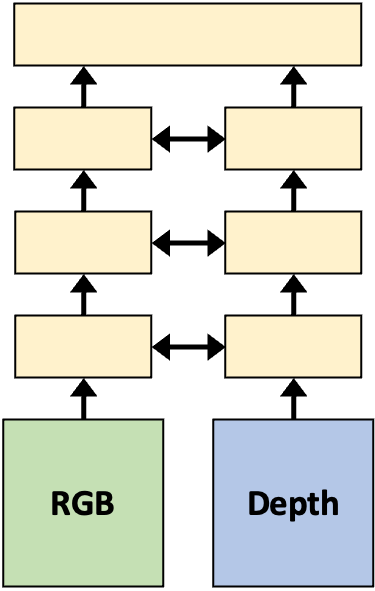}}\hfill
    \subfloat[Intra-inter (Ours)]{\includegraphics[width=0.2156\linewidth]{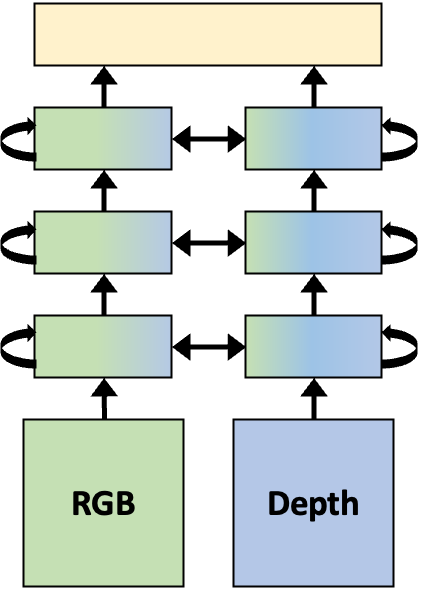}}\hfill
\captionsetup{labelfont={small,bf}, textfont=small}
\caption{Overview of baseline models with the proposed method for RGB-D instance segmentation.}
\label{Fig:Figure5}
\end{figure}

We first introduce a variety of baseline models for evaluation on the RGB-D instance segmentation dataset we constructed, as depicted in Figure~\ref{Fig:Figure5}. These models include distinct fusion approaches: ``Early fusion'', ``Late fusion'', ``Intra-modal attention'', and ``Inter-modal attention''. The early fusion approach merges RGB and depth data at the start, whereas late fusion processes them through separate networks before merging the outcomes at the end. On the other hand, intra-modal attention delves into the relationships within each modality, applying self-attention throughout its process, while inter-modal attention also processes modalities separately but focuses on how they interact with each other.
However, these approaches have drawbacks: early and late fusion might overlook detailed interactions by merging data only once, and intra-modal and inter-modal attention is limited to focusing exclusively on within or between modality relationships.
This oversight may ultimately hinder the model's ability to generalize well~\cite{DBLP:conf/icml/DuTLLYWYZ23}.
To overcome these limitations, we present a new fusion module called IAM (Intra-modal Attention Mix) that takes into account both relationships within and between modalities together.

\subsection{Proposed Method}

\subsubsection{Intra-modal Attention Mix (IAM)}

Recent studies~\cite{peng2022balanced,DBLP:conf/icml/DuTLLYWYZ23} indicate that existing multimodal approaches primarily focus on the interactions between different modalities and how they can effectively complement each other, yet they fail to fully utilize the capabilities of individual modalities. This study introduces the Intra-modal Attention Mix (IAM) module, aimed at improving modality-specific attention while minimizing cross-modal interactions to lower the computational costs of fusion methods. The IAM module enhances performance and reduces computational complexity by efficiently using information within each modality.

\begin{figure}[t]
\centerline{\includegraphics[width=0.9\columnwidth]{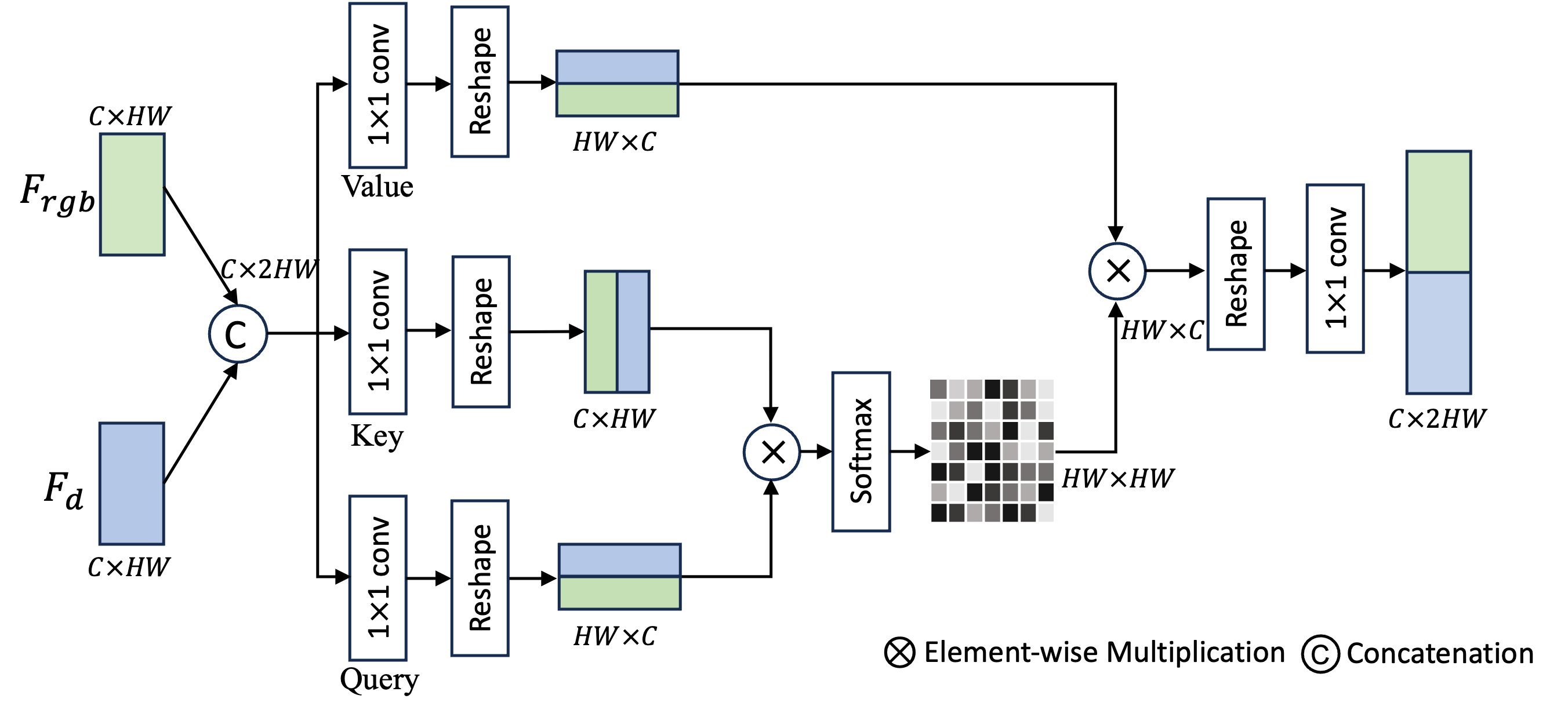}}
\caption{The details of the Intra-modal Attention Mix (IAM) architecture. IAM is a lightweight architecture that focuses on modality-specific interactions and employs an efficient structure that utilizes the interaction between modalities by mixup strategy.}
\label{Fig:Figure6}
\end{figure}

At each step of our process, RGB and depth feature maps denoted as ${F}_{rgb}$ and ${F}_{d}$, are processed using the IAM module. As shown in Figure~\ref{Fig:Figure6}, these maps are described as vectorized RGB and depth feature maps, ${{F}_{rgb}}\in \mathbb{R}^{C\times{H}{W}}$ and ${{F}_{d}}\in \mathbb{R}^{C\times{H}{W}}$, where $H$ and $W$ are spatial dimensions, and $C$ is the number of channels. The combined feature map  ${{F}_{rgbd}}\in \mathbb{R}^{C\times 2HW}$ can be defined as:
\begin{equation}
F_{rgbd}=concat(F_{rgb},F_{d})\in\mathbb{R}^{C\times 2HW}. 
\end{equation}
{$1 \times 1$ convolution (also known as pointwise convolution) is then applied to the combined feature maps. This convolution reduces the channel dimension by half, resulting in a feature map of size $\mathbb{R}^{\frac{C}{2} \times 2HW}$. This step efficiently compresses the feature representation while preserving spatial relationships, thereby decreasing computational complexity for subsequent processes. The reduced feature map is then reshaped to derive Query, Key, and Value vectors as follows:}
\begin{equation}
Q=Reshape(Conv_{1\times 1}^{Q}(F_{rgbd})) \in \mathbb{R}^{HW\times C},
\end{equation}
\begin{equation}
K=Reshape(Conv_{1\times 1}^{K}(F_{rgbd})) \in \mathbb{R}^{HW\times C},
\end{equation}
\begin{equation}
V=Reshape(Conv_{1\times 1}^{V}(F_{rgbd})) \in \mathbb{R}^{HW\times C}.
\end{equation}

After the reshaping process, RGB and Depth information are still clearly distinguished, which is evident from Figure~\ref{Fig:Figure6}. 

At the core of our model is the generation of enhanced intra-modality attention maps of size ${\mathbb{R}^{{HW}\times{HW}}}$. This is achieved by multiplying Query and Key vectors that include both RGB and depth features. {For clearer explanation, we can decompose Query and Key into block matrix form, where $Q_{rgb}, Q_d \in \mathbb{R}^{{HW} \times \frac{C}{2}}$ and $K_{rgb}, K_d \in \mathbb{R}^{{HW} \times \frac{C}{2}}$, representing the RGB and depth features, respectively:}
\begin{equation}
Q = \begin{bmatrix} Q_{rgb} & Q_d \end{bmatrix},
\end{equation}
\begin{equation}
K = \begin{bmatrix} K_{rgb} & K_d \end{bmatrix}.
\end{equation}

Using these block matrices, the attention score calculation focuses on the intra-modal interactions:
\begin{equation}
QK^T = \begin{bmatrix} Q_{rgb} & Q_d \end{bmatrix} \begin{bmatrix} K_{rgb}^T \\ K_d^T \end{bmatrix} = Q_{rgb}K_{rgb}^T + Q_dK_d^T,
\end{equation}
which separates the interactions within the RGB and depth modalities without cross-modal terms. Note that each row in Q and K represents a particular feature channel across the spatial dimension $HW$ (flattened grid of height and width) and each column represents the activation's contribution from a specific spatial location to forming the attention scores.
When we compute the dot product, $Q_{rgb}K_{rgb}^{T}$ yields an attention map focused purely on the relationships within the RGB features and $Q_{d}K_{d}^{T}$ gives an attention map capturing interactions within the depth features. 
{Through summation, the relationships within each modality are established, inspired by the mixup-like~\cite{zhang2018mixup} concept of blending features across modalities. The resulting attention map is calculated as:}
\begin{equation} \begin{aligned} Attention_{map} &= Softmax\left( \frac{QK^T}{\sqrt{d_k}} \right) \\ &= Softmax\left( \frac{Q_{rgb}K_{rgb}^T + Q_dK_d^T}{\sqrt{d_k}} \right) \end{aligned} \end{equation}

{This attention map highlights how each spatial position interacts within the same modality, capturing meaningful relationships without introducing unnecessary complexity. Since RGB and depth data are pixel-aligned, their interactions can be modeled effectively using simple element-wise addition. This alignment ensures computational efficiency while preserving modality-specific information.
Finally, the attention map is combined with the Value vector to project the refined features back to their original spatial dimensions:}
\begin{equation}
Z=Conv(Softmax(QK^T)V) \in \mathbb{R}^{{C}\times{2HW}}. \label{1}
\end{equation}
In the initial stages of our IAM process, we combined RGB and depth data to create the integrated feature $Z$, ensuring that within the combined feature $Z$, the unique characteristics of each data type are maintained and clarified by recognizing the importance of each data type.

\begin{figure}[!t]
\centerline{\includegraphics[width=0.9\columnwidth]{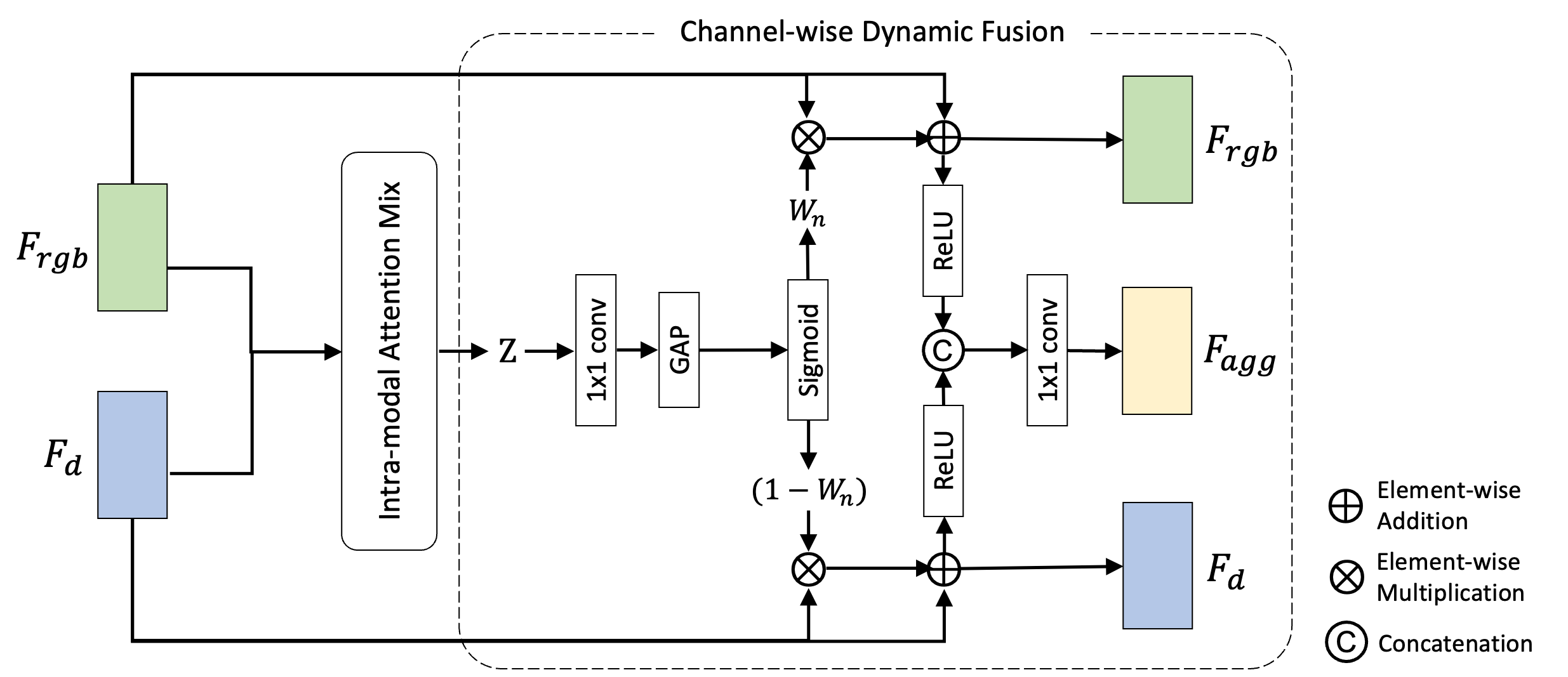}}
\caption{Channel-wise dynamic fusion architecture. It dynamically refines feature interactions within channels, while IAM emphasizes spatial connections.}
\label{Fig:Figure7}
\end{figure}

\subsubsection{Channel-wise Dynamic Fusion}

Expanding on the IAM module, which effectively combines modalities by focusing on spatial characteristics, we introduce the channel-wise dynamic fusion module. This module extends fusion to the channel level, enabling learnable calibration across both spatial and channel dimensions. This ensures a more comprehensive integration of each modality, leading to enhanced performance in RGB-D segmentation.

As illustrated in Figure~\ref{Fig:Figure7}, the module takes the fused feature $Z\in\mathbb{R}^{C\times 2HW}$, the output of the IAM module, as input. 
The fused feature undergoes a series of transformations to dynamically calibrate channel-wise importance. First, a 
$1\times 1$ convolution layer is applied to refine the features, followed by a global average pooling (GAP) layer that summarizes spatial information for each channel. The resulting vector is then normalized using a sigmoid function to produce interpretable weights, represented as: 
\begin{equation}
{W}_{n}=\sigma(GAP(Conv(Z))), \label{2}
\end{equation}
The derived weight vector ${W}_{n}$ represents the relative importance at the channel level between RGB and depth modalities. This aids in making better fusion decisions.
Using this information, we obtain enhanced RGB feature and depth features to propagate to the next layer:
\begin{equation}
{F}_{rgb}=({W}_{n}\otimes{F}_{rgb}+F_{rgb})/2 \in \mathbb{R}^{{C}\times{HW}}, \label{3}
\end{equation}
\begin{equation}
{F}_{d}=((1-{W}_{n})\otimes{F}_{d}+F_{d})/2\in \mathbb{R}^{{C}\times{HW}}. \label{4}
\end{equation}

Subsequently, the enhanced modality-specific features passed through ReLU activation function, concatenated along the channel dimension, and processed through a convolutional layer to obtain the final fused feature map ${{F}_{agg}}$:
\begin{equation}
{F}_{agg}= Conv(Concat(ReLU({F}_{rgb})+ReLU({F}_{d})))\in \mathbb{R}^{{C}\times{HW}}. \label{5}
\end{equation}

The Channel-wise Dynamic Fusion module dynamically calibrates channel importance, outperforming traditional methods that assign equal weights. It enhances both spatial and channel-level integration, improving segmentation performance. Unless specified, the IAM module includes this extension, further refining feature fusion for robust segmentation.

\subsection{Model Architecture}

\begin{figure}[]
\centerline{\includegraphics[width=0.8\columnwidth]{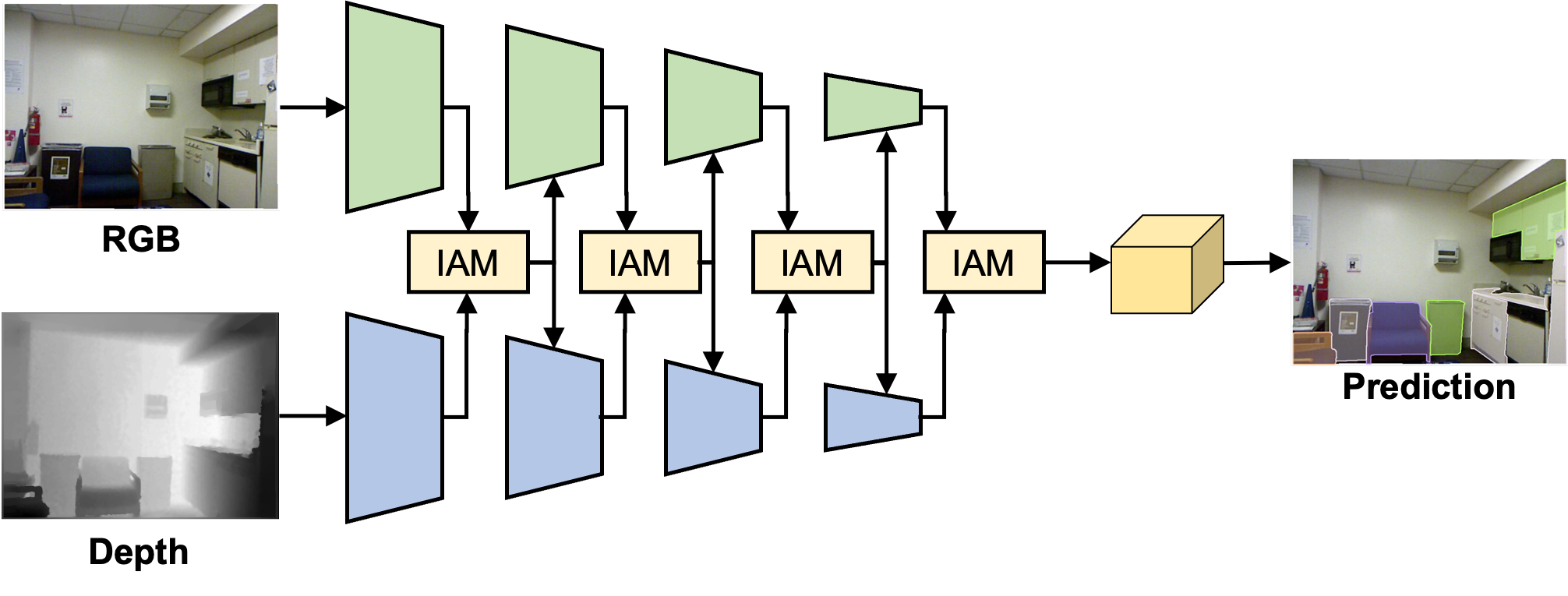}}
\caption{Structure of the adaptable IAM module within dual-stream architectures for RGB-D segmentation.}
\label{fig:Figure12}
\end{figure}

\begin{figure*}[]
\centerline{\includegraphics[width=0.9\textwidth]{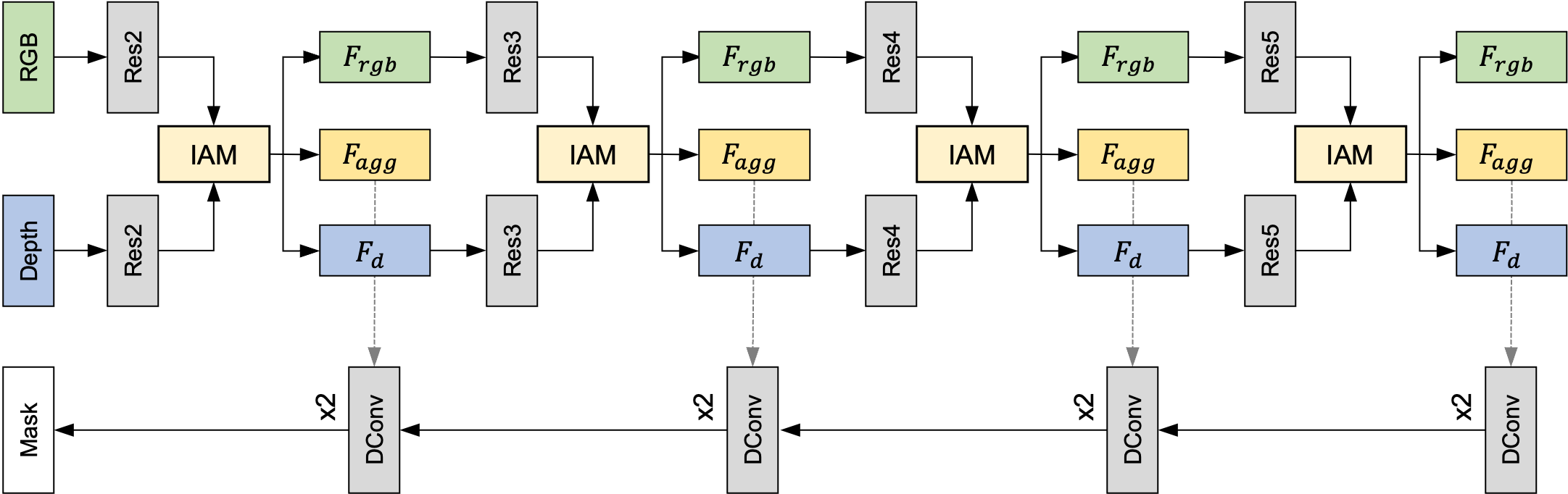}}
\caption{The network structure reconfigured to incorporate the proposed method into the DETR architecture.}
\label{fig:Figure13}
\end{figure*}

We extended the original DETR~\cite{carion2020end} and SOLQ~\cite{dong2021solq} architectures by incorporating a depth stream, as illustrated in Figure~\ref{fig:Figure12}. This depth stream operates in parallel to the RGB branch, using an identical network structure that is trained simultaneously.
Figure~\ref{fig:Figure13} depicts the reconfigured network architecture, enhanced with the proposed IAM module. The IAM module generates three key features: enhanced RGB features (${F}{rgb}$), aggregated RGB-depth features (${F}{agg}$), and enhanced depth features (${F}{d}$). Among these, ${F}{rgb}$ and ${F}{d}$ are forwarded to the subsequent network layers. Meanwhile, ${F}{agg}$ is utilized in the deconvolution layer~\cite{zeiler2010deconvolutional} for segmentation and simultaneously passed to a separate branch for detection tasks.
SOLQ employs a comparable structure but substitutes the deconvolution layers with a unified query representation, enabling joint detection and segmentation. In contrast, DETR separates these tasks, handling detection and segmentation through distinct processing steps.

We further explored various design options regarding the placement of ${F}{rgb}$ and ${F}{agg}$. Among these, the configuration where ${F}{rgb}$ is utilized for detection and ${F}{agg}$ for segmentation demonstrated the most effective performance. Consequently, this design was chosen for DETR-based instance segmentation. Detailed analyses and results will be presented in the experimental section.

\section{Experiments}

\subsection{Implementation Details}
\label{ssec:1}

Within the two-stream architecture, the IAM block is inserted between the corresponding layers of the two modalities, where inputs encompass RGB and depth images. For RGB-D instance segmentation, experiments were conducted using the DETR~\cite{carion2020end} and SOLQ~\cite{dong2021solq} models to assess the impact of the IAM module. DETR handles detection and segmentation in separate branches, whereas SOLQ processes them in a unified branch.

Training was performed with a batch size of 4 across all experiments. SOLQ was trained for 50 epochs with an initial learning rate of $2.0 \times 10^{-4}$, reduced by a factor of 0.1 every 15 epochs. DETR was trained for 50 epochs with an initial learning rate of $1.0 \times 10^{-4}$, also reduced by a factor of 0.1 every 15 epochs. For the Box-IS dataset, DETR training was extended to 200 epochs, with the learning rate decayed at epoch 130. All other hyperparameters followed the default configurations provided in the original DETR and SOLQ implementations.

\subsection{Dataset Split}
\label{ssec:2}

For model training and evaluation, the datasets were split as follows:
\begin{itemize}
    \item \textbf{NYUDv2-IS}: 788 training samples and 645 validation samples, following the configuration in prior semantic segmentation research~\cite{xie2021segformer}.
    \item \textbf{SUN-RGBD-IS}: 5,070 training samples and 4,872 validation samples.
    \item \textbf{Box-IS}: 488 training samples and 55 validation samples.
    \item \textbf{SUN-RGBD-IS-kv2}: Built using only Kinect v2 data from SUN-RGBD-IS, containing 2,838 training samples and 946 validation samples. This dataset was constructed to evaluate the IAM module under better depth quality conditions compared to Kinect v1.
\end{itemize}

\subsection{Comparison with Previous Fusion Methods}
\label{ssec:3}

We conducted RGB-D instance segmentation experiments on DETR and SOLQ to evaluate the IAM module. Our method was compared with  ``Early fusion'', ``Late fusion'',``Intra'', and ``Inter'' methods, as well as state-of-the-art modules like SA-Gate~\cite{chen2020bi} and CMX~\cite{zhang2022cmx}. Additionally, experiments were conducted across datasets (NYUDv2-IS, SUN-RGBD-IS, and Box-IS) to assess generalization.

In all settings, models using our IAM module consistently outperformed RGB-only models. As shown in Table~\ref{table:3}, integrating IAM into DETR and SOLQ resulted in significant performance improvements, with gains of 6.7\% and 7.2\% $\mathrm{AP}^{seg}$ over early fusion, respectively. Similarly, our method surpassed SA-Gate and CMX by margins of 1.6\% and 0.9\% (DETR) and 0.4\% and 3.3\% (SOLQ) in $\mathrm{AP}^{seg}$. Furthermore, IAM demonstrated superiority over intra- and inter-modal attention modules.
Table~\ref{table:4} and Table~\ref{table:5} further illustrate the effectiveness of IAM across datasets. While SOLQ consistently showed better performance with IAM, DETR's architecture—which separates detection and segmentation—posed challenges for leveraging IAM's integrated representations, particularly on simpler datasets like Box-IS. Despite this, IAM demonstrated notable performance enhancements across tasks, proving its adaptability.

\setlength{\tabcolsep}{0.12cm}
\begin{table}[!t]
\begin{center}

\footnotesize
\resizebox{0.7\columnwidth}{!}{
\begin{tabular}{l|l|cc|ccccc}
\toprule
\multicolumn{2}{c|}{Method} & $\mathrm{AP}^{seg}$ & $\mathrm{AP}^{det}$ & ${\mathrm{AP}_{0.5}^{seg}}$ & ${\mathrm{AP}_{0.75}^{seg}}$ & ${\mathrm{AP}_{S}^{seg}}$ & ${\mathrm{AP}_{M}^{seg}}$ & ${\mathrm{AP}_{L}^{seg}}$ \\
\midrule
\multirow{7}{*}{\rotatebox{90}{\footnotesize DETR}}
& RGB & 31.2 & 39.0 & 53.8 & 31.7 & 5.3 & 18.9 & 41.1 \\
& Early & 25.6 & 32.8 & 45.4 & 25.9 & 4.1 & 14.6 & 34.3 \\
& Late & 31.9 & 38.8 & 53.9 & 32.5 & 5.2 & 19.5 & 41.7 \\
& Intra & 29.5 & 37.9 & 51.6 & 29.4 & 5.7 & 18.2 & 38.9 \\
& Inter & 30.1 & 38.3 & 52.3 & 30.4 & 3.8 & 19.0 & 39.4 \\
& SA-Gate & 30.7 & 38.4 & 52.5 & 31.2 & 4.2 & 19.2 & 40.2 \\
& CMX & 31.4 & 39.5 & 54.2 & 31.9 & 4.6 & \textbf{20.7} & 40.4 \\
& Ours  & \textbf{32.3} & \textbf{39.8} & \textbf{54.3} & \textbf{32.9} & \textbf{6.2} & 20.0 & \textbf{42.2} \\
\midrule
\multirow{7}{*}{\rotatebox{90}{\footnotesize SOLQ}}
& RGB  & 33.1 & 40.5 & 52.8 & 34.7 & 3.2 & 20.6 & 44.5 \\
& Early  & 28.6 & 36.6 & 47.8 & 29.9 & 2.5 & 16.3 & 38.9 \\
& Late  & 34.9 & 42.1 & 55.5 & 37.5 & \textbf{5.6} & 21.5 & \textbf{46.5} \\
& Intra & 35.3 & 42.8 & 56.0 & 37.6 & 3.6 & 23.9 & 46.9 \\
& Inter & 35.1 & 42.8 & 55.5 & 37.4 & 4.5 & 23.5 & 45.6 \\
& SA-Gate & 35.4 & 42.9 & 56.1 & 38.1 & 3.8 & 25.1 & 46.2 \\
& CMX & 32.5 & 39.8 & 35.3 & 34.6 & 4.6 & 20.4 & 43.4 \\
& Ours  & \textbf{35.8} & \textbf{43.7} & \textbf{57.1} & \textbf{38.4} & 3.7 & \textbf{26.2} & 46.0 \\
\bottomrule
\end{tabular}
}
\end{center}
\vspace{-1em}
\caption{Performance comparison of IAM using various methods for RGB-D instance segmentation on NYUDv2-IS.
}
\label{table:3}
\end{table}
\setlength{\tabcolsep}{0.12cm}

\setlength{\tabcolsep}{0.12cm}
\begin{table}[!t]
\begin{center}
\footnotesize
\resizebox{0.7\columnwidth}{!}{
\begin{tabular}{l|l|cc|ccccc}
\toprule
\multicolumn{2}{c|}{Method} & $\mathrm{AP}^{seg}$ & $\mathrm{AP}^{det}$ & ${\mathrm{AP}_{0.5}^{seg}}$ & ${\mathrm{AP}_{0.75}^{seg}}$ & ${\mathrm{AP}_{S}^{seg}}$ & ${\mathrm{AP}_{M}^{seg}}$ & ${\mathrm{AP}_{L}^{seg}}$ \\
\midrule
\multirow{6}{*}{\rotatebox{90}{\footnotesize DETR}}
& RGB & 22.5 & 28.8 & 39.3 & 22.7 & 1.4 & 11.1 & 29.3 \\
& Early  & 20.6 & 26.4 & 35.9 & 20.5 & 0.9 & 8.9 & 27.0 \\
& Late   & 22.5 & 28.7 & 38.4 & 22.8 & 2.2 & 11.1 & 29.0 \\
& Intra & \textbf{23.6} & \textbf{30.2} & \textbf{40.1} & \textbf{23.9} & 2.8 & \textbf{11.5} & \textbf{30.6} \\
& Inter & 23.4 & 30.1 & 40.0 & 23.6 & \textbf{3.3} &  11.2 & 30.3 \\
& Ours & 22.9 & 29.3 & 39.5 & 22.7 & 2.5 & 11.3 & 29.2 \\
\midrule
\multirow{6}{*}{\rotatebox{90}{\footnotesize SOLQ}}
& RGB & 23.6 & 29.2 & 38.7 & 24.2 & 2.7 & 11.7 & 30.2 \\
& Early  & 23.4 & 29.3 & 38.2 & 24.2 & 2.0 & 10.7 & 30.2 \\
& Late & 24.2 & 30.1 & 39.5 & 25.2 & 3.6 & 11.4 & 31.3 \\
& Intra & 24.4 & 30.9 & 39.8 & 25.3 & 4.9 & 11.9 & 31.2   \\
& Inter & 24.7 & 30.8 & 40.3 & 25.8 & 4.0 & 11.5 & 31.9   \\
& Ours & \textbf{25.7} & \textbf{31.7} & \textbf{41.3} & \textbf{26.8} & \textbf{5.9} & \textbf{12.0} & \textbf{33.1} \\
\bottomrule
\end{tabular}
}
\end{center}
\vspace{-1em}
\caption{Performance comparison of IAM using various methods for RGB-D instance segmentation on SUN-RGBD-IS.}
\label{table:4}

\end{table}
\setlength{\tabcolsep}{0.12cm}

\setlength{\tabcolsep}{0.12cm}
\begin{table}[!t]
\begin{center}
\footnotesize
\resizebox{0.7\columnwidth}{!}{
\begin{tabular}{l|l|cc|ccccc}
\toprule
\multicolumn{2}{c|}{Method} & $\mathrm{AP}^{seg}$ & $\mathrm{AP}^{det}$ & ${\mathrm{AP}_{0.5}^{seg}}$ & ${\mathrm{AP}_{0.75}^{seg}}$ & ${\mathrm{AP}_{S}^{seg}}$ & ${\mathrm{AP}_{M}^{seg}}$ & ${\mathrm{AP}_{L}^{seg}}$ \\

\midrule
\multirow{6}{*}{\rotatebox{90}{\footnotesize SOLQ}}
& RGB  & 83.3 & 89.0 & 93.5 & 87.6 & 7.9 & 54.0 & 87.9 \\
& Early  & 81.6 & 87.9 & 92.6 & 86.4 & 1.8 & 52.0 & 86.9 \\
& Late  & 83.2 & 89.5 & 92.9 & 87.0 & \textbf{31.5} & 55.1 & 87.8 \\
& Intra & 83.4 & 89.6 & \textbf{94.6} & \textbf{87.7} & 14.0 & \textbf{56.3} & 87.8   \\
& Inter & 83.3 & \textbf{89.9} & 93.2 & 87.1 & 14.6 & 55.2 & 88.0   \\
& Ours & \textbf{83.7} & 89.2 & 93.7 & 87.5 & 7.2 & 52.7 & \textbf{88.4} \\
\bottomrule
\end{tabular}
}
\end{center}
\vspace{-1em}
\caption{Performance comparison of IAM using various methods for RGB-D instance segmentation on Box-IS.}
\label{table:5}
\end{table}
\setlength{\tabcolsep}{0.12cm}

\subsection{Ablation Study}
\label{ssec:4}
{To evaluate the effectiveness of our method, we conducted comprehensive ablation experiments on instance segmentation on the NYUDv2-IS dataset.}

\setlength{\tabcolsep}{0.2cm}
\begin{table}[!t]
\begin{center}
\scriptsize
\resizebox{0.55\columnwidth}{!}{%
\begin{tabular}{lccccccc}
\toprule
Backbone  & Block2 & Block3 & Block4 & $\mathrm{AP}^{seg}$  \\
\midrule
ResNet50 & & & & 28.6 \\
ResNet50 & \checkmark & & & 34.9\\
ResNet50 & & \checkmark & & 34.9 \\
ResNet50 & & & \checkmark & 34.5 \\
ResNet50 & \checkmark & \checkmark & & 34.7 \\
ResNet50 & \checkmark & & \checkmark & 35.0\\
ResNet50 & & \checkmark & \checkmark & 34.6\\
ResNet50 & \checkmark & \checkmark & \checkmark & \textbf{35.8} \\
\bottomrule
\end{tabular}}
\end{center}
\vspace{-1em}
\caption{Performance comparison based on the block placement for RGB-D instance segmentation using SOLQ on NYUDv2-IS.}
\label{table:6}
\end{table}
\setlength{\tabcolsep}{0.2cm}

\subsubsection{Position of IAM Blocks}
Table~\ref{table:6} highlights the impact of block placement on performance. Incorporating IAM across all possible blocks maximized segmentation accuracy, achieving a 7.2\% $\mathrm{AP}^{seg}$ gain compared to setups without IAM. Constraints prevented using IAM in the first block due to memory limitations.

\begin{figure*}[]
\centering
\captionsetup{labelfont=small, textfont=small}
    \subfloat[Design A]    {\includegraphics[width=0.38\linewidth]{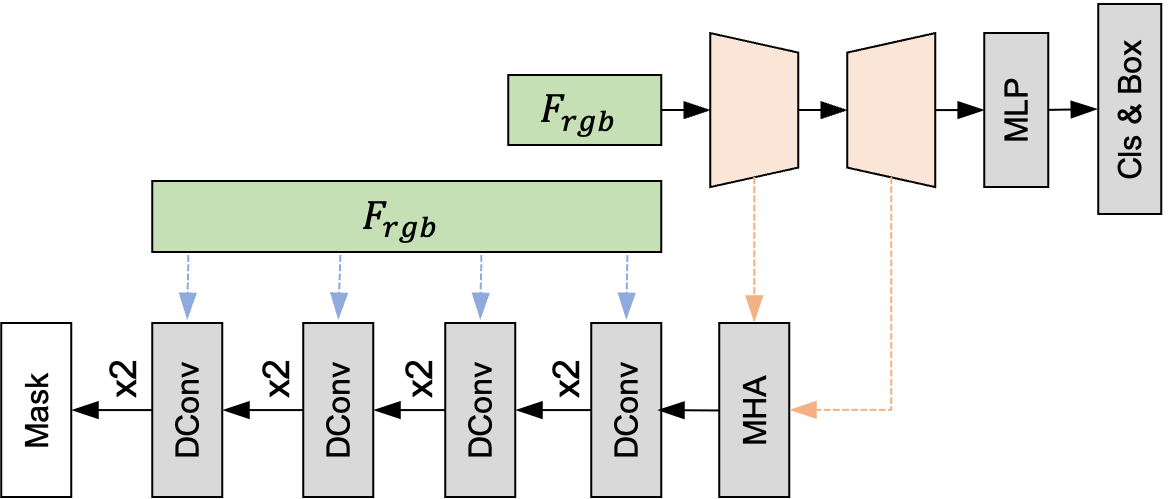}}\hspace{1cm}
    \subfloat[Design B]
{\includegraphics[width=0.38\linewidth]{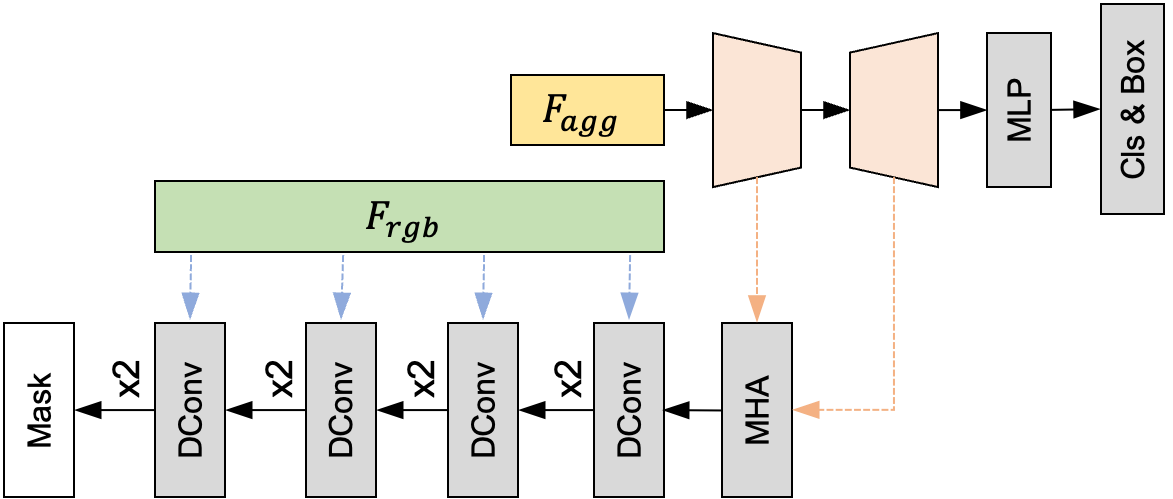}}\hspace{1cm}
    \subfloat[Design C]
{\includegraphics[width=0.38\linewidth]{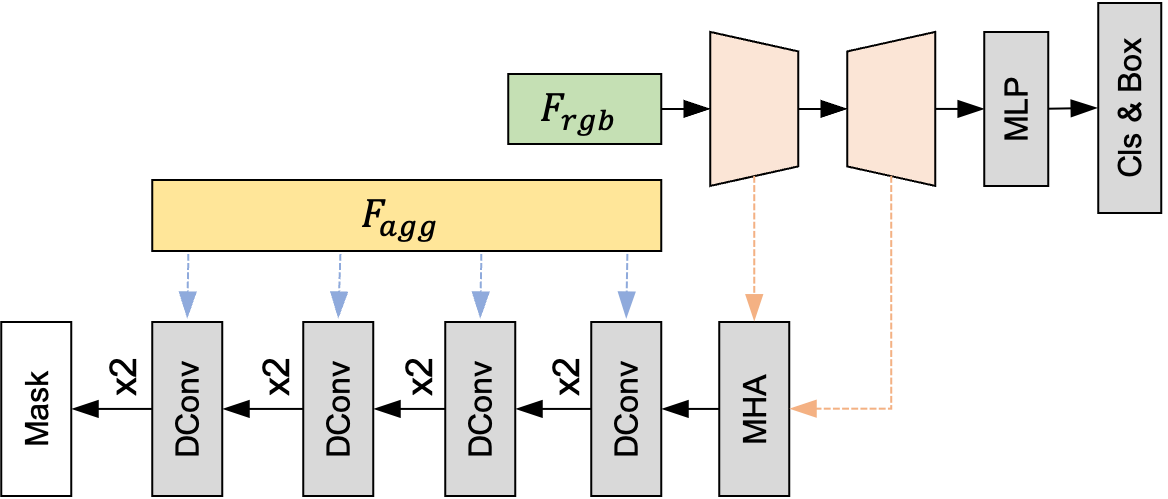}}\hspace{1cm}
    \subfloat[Design D]
{\includegraphics[width=0.38\linewidth]{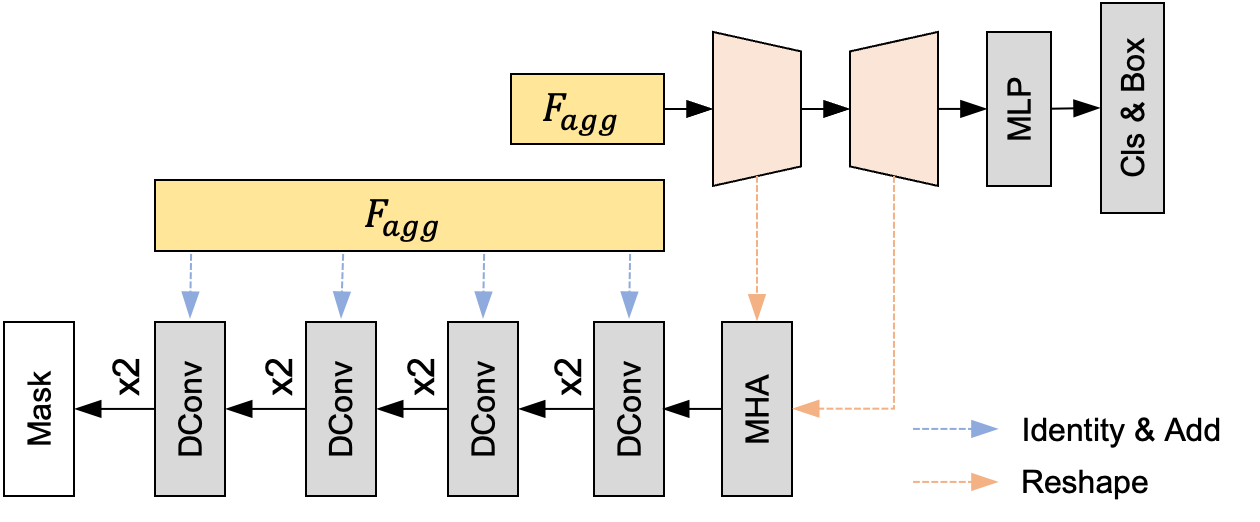}}
\captionsetup{labelfont={small}, textfont=small}
\caption{Comparison of DETR structures that fuse depth with RGB.}
\label{Fig:Figure14}
\end{figure*}

\setlength{\tabcolsep}{0.2cm}
\begin{table}[!th]
\begin{center}

\footnotesize
\resizebox{0.6\columnwidth}{!}{
\begin{tabular}{l|cc|ccccc}
\toprule
Method & $\mathrm{AP}^{seg}$ & $\mathrm{AP}^{det}$ & ${\mathrm{AP}_{S}^{seg}}$ & ${\mathrm{AP}_{M}^{seg}}$ & ${\mathrm{AP}_{L}^{seg}}$ \\
\midrule
Design A & 31.2 & 38.7 & 19.9 & 27.3 & 46.4 \\
Design B & 31.6 & 38.0 & 18.1 & 27.9 & 46.0 \\
Design C  & \textbf{32.3} & \textbf{39.8} & \textbf{22.2} & \textbf{28.7} & \textbf{47.6} \\
Design D & 31.2 & 37.6 & 15.0 & 28.1 & 45.4 \\
\bottomrule
\end{tabular}
}
\end{center}
\vspace{-1em}
\caption{DETR-based design choice performance comparison on NYUDv2-IS.}
\label{table:1s}
\end{table}
\setlength{\tabcolsep}{0.2cm}

\subsubsection{Design Choices for Feature Integration}
Figure~\ref{Fig:Figure14} illustrates the comparison of DETR structures that fuse depth with RGB. This figure provides a detailed explanation of the different design settings, as described in the text. Specifically, it supports the evaluation mentioned in Table~\ref{table:1s}, where Design C outperformed other configurations by utilizing ${F}{rgb}$ for detection and ${F}{agg}$ for segmentation, achieving the highest $\mathrm{AP}^{seg}$ and $\mathrm{AP}^{det}$ metrics with balanced RGB-depth utilization.

\subsubsection{Effect of Each Module}
As detailed in Table~\ref{table:7}, combining IAM and CDF modules resulted in a substantial boost of 7.2\% $\mathrm{AP}^{seg}$. Individually, IAM and CDF contributed 5.8\% and 4.7\%, respectively, further validating their effectiveness. The consistent gains demonstrate the complementary nature of these modules, making them valuable additions to RGB-D instance segmentation pipelines.

\subsubsection{Effect of Sensor Quality}
Table~\ref{table:2s} shows how improved depth quality from SUN-RGBD-IS-kv2 enhanced overall segmentation and detection. IAM consistently delivered the best performance regardless of sensor type, underscoring its adaptability. These findings highlight that our approach is resilient to variations in data quality.

\setlength{\tabcolsep}{0.2cm}
\begin{table}[t]
\begin{center}
\footnotesize
\resizebox{0.55\columnwidth}{!}{%
\begin{tabular}{lccccc}
\toprule
Method & $\mathrm{AP}^{seg}$(\%) & $\mathrm{AP}^{det}$(\%) \\
\midrule
SOLQ & 28.6 & 36.6 \\
SOLQ + CDF & 33.3 & 41.3 \\
SOLQ + IAM & 34.4 & 42.0 \\
SOLQ + IAM + CDF & \textbf{35.8} & \textbf{43.7} \\
\bottomrule
\multicolumn{2}{l}{{CDF}: Channel-wise Dynamic Fusion}
\end{tabular}}
\end{center}
\vspace{-1em}
\caption{Ablation study on the effect of each module for RGB-D instance segmentation using SOLQ on NYUDv2-IS.}
\label{table:7}
\end{table}
\setlength{\tabcolsep}{0.2cm}

\setlength{\tabcolsep}{0.12cm}
\begin{table}[!t]
\begin{center}
\footnotesize
\resizebox{0.7\columnwidth}{!}{
\begin{tabular}{l|l|cc|ccccc}
\toprule
\multicolumn{2}{c|}{Method} & $\mathrm{AP}^{seg}$ & $\mathrm{AP}^{det}$ & ${\mathrm{AP}_{0.5}^{seg}}$ & ${\mathrm{AP}_{0.75}^{seg}}$ & ${\mathrm{AP}_{S}^{seg}}$ & ${\mathrm{AP}_{M}^{seg}}$ & ${\mathrm{AP}_{L}^{seg}}$ \\
\midrule
\multirow{8}{*}{\rotatebox{90}{\footnotesize DETR}}
& RGB & 19.0 & 23.3 & 32.7 & 19.6 & 1.1 & 11.7 & 25.6 \\
& Early  & 17.0 & 21.8 & 30.2 & 16.5 & 1.1 & 10.8 & 23.7 \\
& Late   & 19.6 & 23.5 & 33.2 & 20.9 & 1.1 & 12.2 & 26.6 \\
& Intra & 18.7 & 23.9 & 32.5 & 19.2 & 1.1 & 9.4 & 27.1 \\
& Inter & 19.4 & 25.1 & 33.0 & 19.9 & 0.6 & 7.8 & 28.5 \\
& SA-Gate & 20.3 & 24.1 & 33.4 & 20.7 & \textbf{1.7} & 8.6 & 28.6 \\
& CMX & 21.0 & 24.7 & 35.1 & \textbf{23.1} & 1.3 & 12.0 & 29.1 \\
& Ours & \textbf{21.1} & \textbf{25.2} & \textbf{35.2} & 21.9 & 0.6 & \textbf{14.7} & \textbf{29.3} \\
\midrule
\multirow{8}{*}{\rotatebox{90}{\footnotesize SOLQ}}
& RGB & 20.4 & 25.1 & 31.5 & 20.8 & 1.1 & 9.7 & 28.6 \\
& Early & 19.4 & 24.4 & 30.2 & 19.9 & 1.0 & 8.4 & 27.3 \\
& Late  & 21.5 & 26.1 & 33.0 & 22.4 & 1.3 & 10.5 & 30.0  \\
& Intra & 22.6 & 28.5 & 35.5 & 23.2 & 2.1 & 11.8 & 31.0      \\
& Inter & 22.4 & 27.2 & 34.2 & 23.5 & 2.1 & 11.3 & 30.9  \\
& SA-Gate & 21.6 & 26.2 & 33.6 & 22.9 & 1.2 & \textbf{12.6} & 29.8\\
& CMX & 18.6 & 23.8 & 30.0 & 19.5 & \textbf{2.8} & 8.1 & 26.6 \\
& Ours & \textbf{23.3} & \textbf{28.5} & 35.4 & \textbf{25.3} & 1.9 & 12.4 & \textbf{31.9} \\
\bottomrule
\end{tabular}
}
\end{center}
\vspace{-1em} %
\caption{Performance comparison of IAM using various methods for RGB-D instance segmentation on SUN-RGBD-IS-kv2.}
\label{table:2s}

\end{table}
\setlength{\tabcolsep}{0.12cm}

\begin{figure*}[]
\centering
\includegraphics[width=1.0\textwidth]{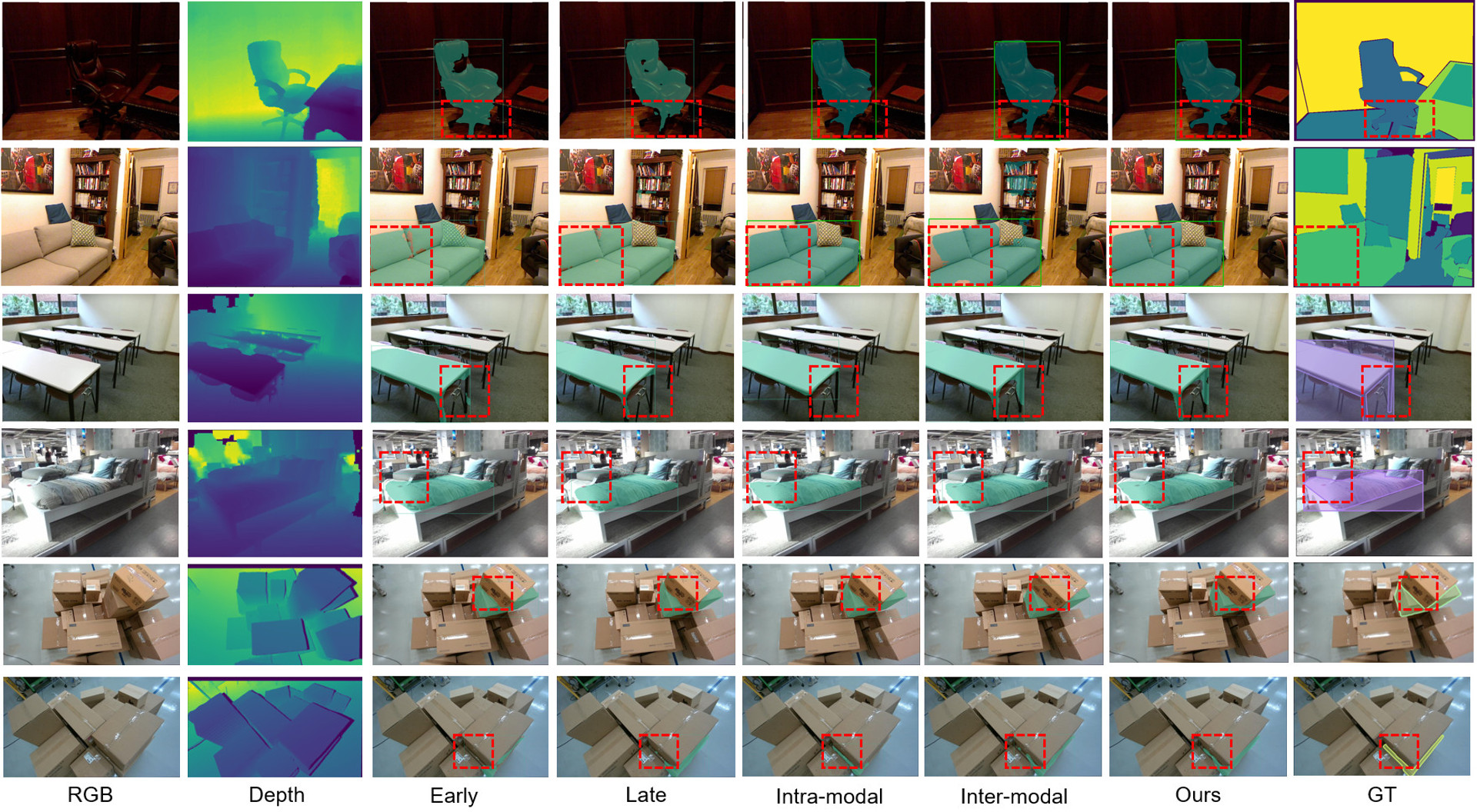}
\caption{A visual comparison of the early fusion, late fusion, intra-modal, inter-modal, and our method on various datasets. It shows their effectiveness in RGB-D instance segmentation. The red boxes highlight where our approach outperforms others. Rows 1-2 are conducted on NYUDv2-IS, rows 3-4 on SUN-RGBD-IS, and rows 5-6 on Box-IS.}
\label{Fig:Figure8}
\end{figure*}

\subsection{Qualitative Analysis}
\label{ssec:5}

Figure \ref{Fig:Figure8} visually illustrates the impact of incorporating the IAM block on instance segmentation performance across the NYUDv2-IS, SUN-RGBD-IS, and Box-IS datasets. Rows 1-2 depict results for NYUDv2-IS, rows 3-4 for SUN-RGBD-IS, and rows 5-6 for Box-IS. The inclusion of IAM blocks results in more refined and uniform segmentation predictions, as highlighted in the side-by-side image comparisons. 

For larger objects and object boundaries, the outputs from intra-modal and inter-modal analyses exhibit noticeable discrepancies compared to the ground truth. In contrast, our approach generates segmentation results that align more closely with the ground truth. Specifically, the IAM module effectively handles complex textures, as demonstrated in the first row by refining the local geometry of depth images and enhancing global connections within RGB images. The fourth row showcases the module's capability to accurately capture table legs, demonstrating its robustness in detecting fine details. Additionally, results from Box-IS reveal that the IAM module significantly improves segmentation accuracy, especially in scenarios involving overlapping objects. Overall, the IAM module enhances object edge delineation and demonstrates its proficiency in utilizing depth information to augment segmentation quality.

\section{Conclusion}
This study highlights the importance of depth information in RGB-D instance segmentation and addresses the scarcity of relevant datasets by introducing three benchmarks for real-world indoor scenarios: NYUDv2-IS, SUN-RGBD-IS, and Box-IS. These datasets fill a critical gap and support applications in indoor navigation, robotics, and assistive systems.
The proposed Intra-modal Attention Mix (IAM) module demonstrated its effectiveness through comprehensive evaluations, enhancing segmentation performance by integrating RGB-D data. Beyond practical contributions, this research provides deeper insights into scene understanding, offering tools and techniques for future advancements.
The broad applicability of this work spans robotics, spatial understanding, and assistive technologies, paving the way for further innovations in RGB-D instance segmentation. We believe this study will inspire ongoing progress in this rapidly evolving field.

\clearpage

\bibliographystyle{elsarticle-num} 
\bibliography{Bibliography.bib}

\end{document}